\documentclass[acmlarge,screen,nonacm]{acmart}

\makeatletter
\def\@ACM@checkaffil{
    \if@ACM@instpresent\else
    \ClassWarningNoLine{\@classname}{No institution present for an affiliation}%
    \fi
    \if@ACM@citypresent\else
    \ClassWarningNoLine{\@classname}{No city present for an affiliation}%
    \fi
    \if@ACM@countrypresent\else
        \ClassWarningNoLine{\@classname}{No country present for an affiliation}%
    \fi
}
\makeatother

\setcopyright{none}
\def\BibTeX{{\rm B\kern-.05em{\sc i\kern-.025em b}\kern-.08emT\kern-.1667em\lower.7ex\hbox{E}\kern-.125emX}}
\renewcommand\footnotetextcopyrightpermission[1]{}
\usepackage{gensymb}
\usepackage{algorithm}

\usepackage{algorithmic}

\AtBeginDocument{%
  \providecommand\BibTeX{{%
    \normalfont B\kern-0.5em{\scshape i\kern-0.25em b}\kern-0.8em\TeX}}}

\begin{document}

\title{WSTac: Interactive Surface Perception based on Whisker-Inspired and Self-Illuminated Vision-Based Tactile Sensor}

\author{Kai~Chong~Lei}
\authornote{Equal Contribution.}
\author{Kit~Wa~Sou}
\authornotemark[1]
\author{Wang~Sing~Chan}
\email{{liqc21, sujh21, chs22}@mails.tsinghua.edu.cn}
\affiliation{%
\institution{Tsinghua-Berkeley Shenzhen Institute}
 \institution{Shenzhen International Graduate School}
  \institution{Tsinghua University}
}

\author{Jiayi~Yan}
\email{yanjy21@mails.tsinghua.edu.cn}
\affiliation{%
 \institution{Shenzhen International Graduate School}
  \institution{Tsinghua University}
}

\author{Siqi~Ping}
\email{psq22@mails.tsinghua.edu.cn}
\affiliation{%
\institution{Tsinghua-Berkeley Shenzhen Institute}
 \institution{Shenzhen International Graduate School}
  \institution{Tsinghua University}
}

\author{Dengfeng~Peng}
\email{pengdengfeng@szu.edu.cn}
\affiliation{%
\institution{Key Laboratory of Optoelectronic Devices and Systems of Ministry of Education and Guangdong Province}
 \institution{College of Physics and Optoelectronic Engineering}
  \institution{Shenzhen University}
}

\author{Wenbo~Ding}
\authornote{Wenbo Ding is the corresponding author. \\
This work is supported by the grant from the National Natural Science Foundation of China under Grants 62104125 and 62311530102, 
Guangdong Innovative and Entrepreneurial Research Team Program (2021ZT09L197), 
Tsinghua Shenzhen InternationalGraduate School-Shenzhen Pengrui Young Faculty Program of Shenzhen Pengrui Foundation(No. SZPR2023005), 
Shenzhen Science and Technology Program (JCYJ20220530143013030).}
\email{ding.wenbo@sz.tsinghua.edu.cn}
\affiliation{%
\institution{Tsinghua-Berkeley Shenzhen Institute}
 \institution{Shenzhen International Graduate School}
  \institution{Tsinghua University}
  \institution{RISC-V International Open Source Laboratory}
}

\author{Xiao-Ping~Zhang}
\email{xiaoping.zhang@sz.tsinghua.edu.cn}
\affiliation{%
\institution{Tsinghua-Berkeley Shenzhen Institute}
\institution{Shenzhen International Graduate School}
  \institution{Tsinghua University}
\institution{Department of Electrical, Computer and Biomedical Engineering}
  \institution{Ryerson University}}
  
\renewcommand{\shortauthors}{K.~C.~Lei and K.~W.~Sou et al.}

\begin{abstract}

\textbf{Abstract:}

Modern Visual-Based Tactile Sensors (VBTSs) use cost-effective cameras to track elastomer deformation, but struggle with ambient light interference. 
Solutions typically involve using internal LEDs and blocking external light, thus adding complexity. 
Creating a VBTS resistant to ambient light with just a camera and an elastomer remains a challenge. 
In this work, we introduce WStac, a self-illuminating VBTS comprising a mechanoluminescence (ML) whisker elastomer, camera, and 3D printed parts. The ML whisker elastomer, inspired by the touch sensitivity of vibrissae, offers both light isolation and high ML intensity under stress, thereby removing the necessity for additional LED modules. With the incorporation of machine learning, the sensor effectively utilizes the dynamic contact variations of 25 whiskers to successfully perform tasks like speed regression, directional identification, and texture classification.
Videos are available at: \url{https://sites.google.com/view/wstac/}.
\end{abstract}

\maketitle

\section{Introduction}

\begin{figure}[thpb]
  \centering
  \includegraphics[width=1.0\linewidth]{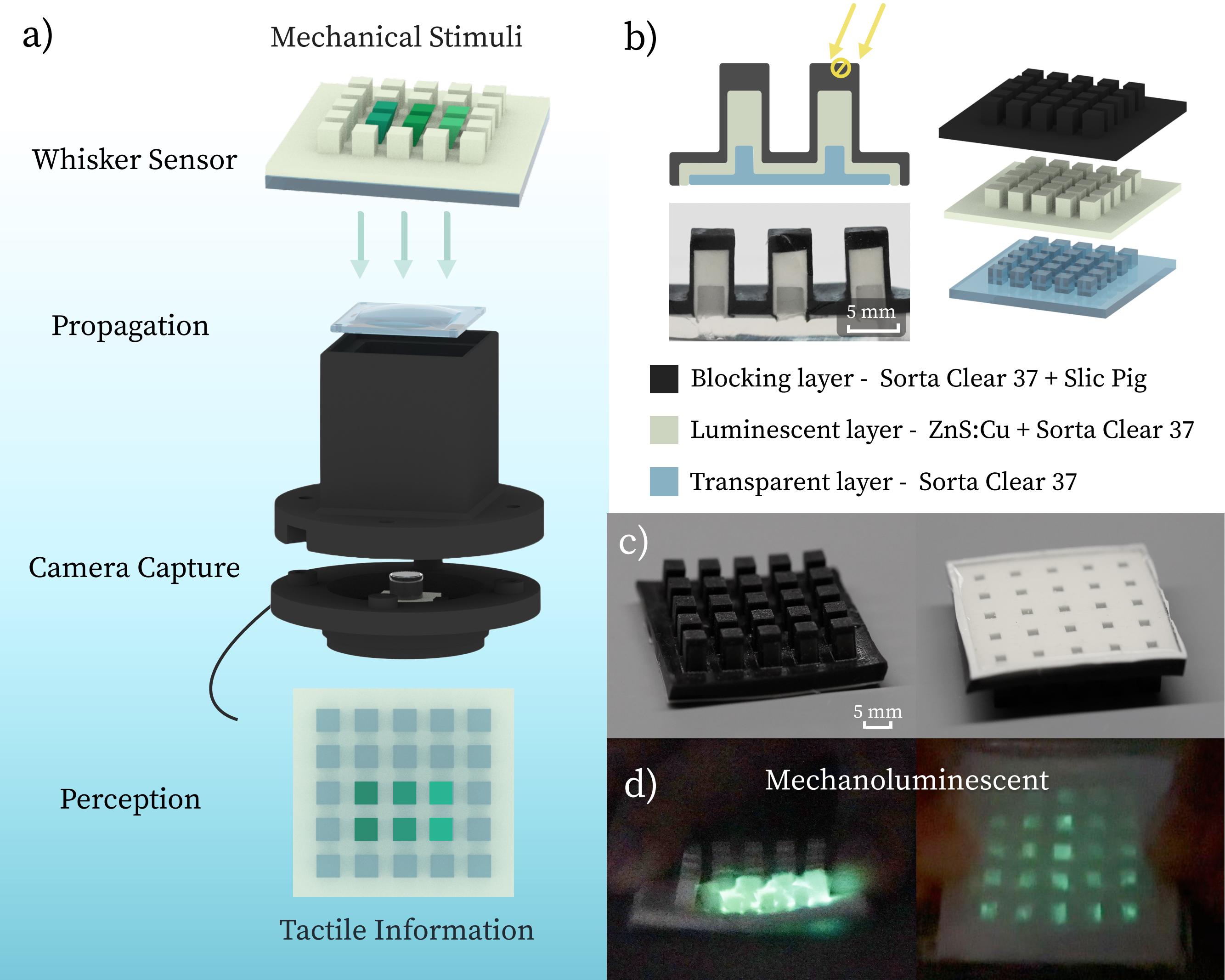}\\
  \caption{(a)  Schematic illustration of the WSTac; (b) Tri-layered elastomer : ML Whisker; (c) ML whisker array; (d) Imaging of ML generated from ML whisker.}\label{conceptDesign}
  \vspace{-6pt}
\end{figure}

As humans and robots interact with the environment, they amass a vast amount of physical information through visual, auditory, and tactile perception~\cite{intro_See_Hear_Feel_CoRL_lifeifei}. In comparison to visual and auditory senses, tactile perception provides a superior recognition of surface characteristics encountered during the interaction, such as geometry ~\cite{intro_tactile_shape_SMC}, temperature~\cite{intro_tactile_thermosensitive_IEEESensorsJournal}, hardness ~\cite{intro_tactile_hardness_IEEESensorsJournal}, materials and texture ~\cite{intro_material_texture_ACM_ubicomp}, etc. To this end, tactile sensors based on different working mechanisms, including piezoresistive ~\cite{intro_piezoresistive_Physical}, resistive~\cite{intro_resistive_IEEESensorsJournal}, capacitive~\cite{intro_capacitive_Nano_Materials}, piezoelectric~\cite{intro_piezoelectric_MaterialsNDesign}, triboelectric~\cite{intro_triboelectric_AMT} and optical~\cite{intro_optical_APR}, have been proposed and applied to improve the intelligence level of human-machine interface~(HMI) devices and robots~\cite{intro_SZU_AIS}. Particularly, a subclass of optical tactile sensors, Vision-Based Tactile Sensors~(VBTSs), can obtain high-resolution tactile information with low-cost cameras to detect the deformation of the elastomer internally~\cite{intro_VBTS1_THUCS__IEEESensorsJournal, intro_VBTS2_IEEESensorsJournal}.

Over the last two decades, research in VBTS has surged. Notable examples like GelForce~\cite{intro_Gelforce_ACM_SIGGRAPH, intro_Gelforce_ACM_CHI}, Gelsight~\cite{intro_Gelsight_CVPR1, intro_Gelsight_sensors}, TacTip ~\cite{intro_TACTIP_family}, have been utilized to distinguish surface properties upon contact~\cite{intro_VBTS_gripper_1, intro_VBTS_gripper_3}.
In the field of robotics, numerous gripper designs incorporating VBTS have emerged~\cite{intro_VBTS_use_1, intro_VBTS_use_2, intro_VBTS_use_3}. When paired with state-of-the-art deep learning algorithms, these grippers enable the execution of complex tasks, including precise control of USB cables~\cite{intro_cable_IJRR}, and oropharyngeal swab sampling~\cite{intro_Oropharyngeal}. 
Furthermore, recently in the domain of HMI, VBTS has also been combined with wearable devices, creating handheld underwater salvage equipment with high-resolution tactile feedback, assisting people in search and grasp activities when vision is not available~\cite{intro_JamTac_SoRo}.

Despite the robustness of VBTS to influences such as temperature, magnetic fields, electric fields, and humidity, susceptibility to ambient light remains a concern as extraneous light sources can affect sensor stability. 
Therefore, recent mainstream VBTS researches mitigate this limitation by blocking external light, and introducing an internal LED lighting module to illuminate the sensor interior. These methods enable the camera to capture elastomer deformation. 
However, the additional module of LED requires extra electronic components and complex wiring, which leads to unnecessary energy consumption\footnote{The power consumption of the Omnivision OVM7692 camera used in the commercial VTBS DIGIT~\cite{intro_digit} is 0.12W.  
Moreover, according to the authors' tests, the power consumption of DIGIT exceeds that of OVM7692 by more than twice, due to the additional energy consumption introduced by LEDs and PCB circuits.} and heat generation. 
Thus, it is an open problem to design an ambient light-resistant VBTS without equipping it with LEDs~\cite{intro_limitation}.

To address the above challenges, we propose WSTac, a selfilluminating VBTS with solely a camera and an elastomer. Specifically, the WSTac integrates a camera, 3D printed parts, and a mechanoluminescence (ML) whisker elastomer which emits light upon experiencing mechanical stress, serving as a key attribute of the design. The ML whisker elastomer derives its inspiration from vibrissae, tactile hairs found in many mammals, recognized for their sensitivity to fluctuating contact variations. 
The ML whisker elastomer is a tri-layered composite with three distinct functions: external light isolation, emission of light upon deformation (ML), and light guidance towards the sensor. 
As shown in Fig.~\ref{conceptDesign}a, WSTac employs a whisker array and a camera to obtain dynamic tactile data, without requiring LED modules, due to the elastomer's ML upon deformation. 
WSTac's feature extraction algorithm compresses the tactile image into a ten-channel time-series signal, detailing the dynamic contact variations of 25 whiskers in real-time~\cite{intro_vibrissae}. 
Its hardware and algorithm performance has been verified via classical tactile tasks.
When paired with fundamental machine learning algorithms, WSTac successfully recognizes sliding direction, sliding speed, texture pattern and texture depth. 
The contributions of our work can be delineated across four dimensions:

\begin{itemize}
    \item A novel sensor hardware design for VBTS, using only a camera and elastomer, immune to ambient light interference. 
    \item  Bionics-inspired tri-layered whisker array elastomer enables ML for dynamic tactile perception in open environments.
    \item A real-time algorithm with a novel feature extraction technique for VBTS lowers the computational cost by reducing complexity from $\Theta(i \times j)$ to $\Theta(i + j)$.
   \item Qualitative and quantitative results indicate outstanding performance of the WSTac on speed regression, directional identification, and texture classification.
\end{itemize}

\begin{figure}[thpb]
  \centering
  \includegraphics[width=1.0\linewidth]{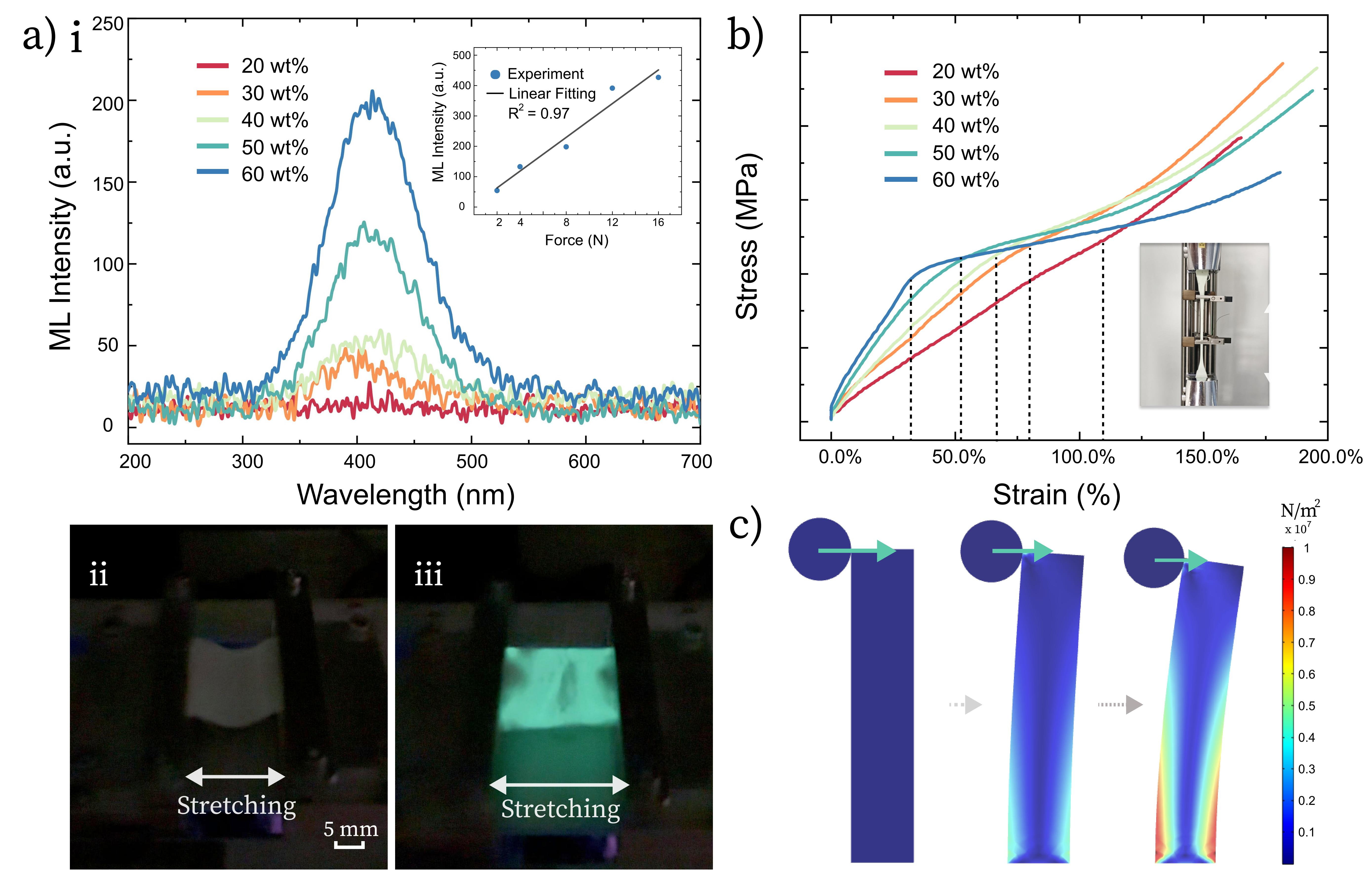}\\
  \caption{(a) i: ML intensity curves with different ZnS:Cu concentrations; ii: Imaging of the ML elastomer brfore stretching; iii: Imaging of the ML elastomer under stretching; (b) Stress–strain curves of ML films with different ZnS:Cu concentrations; (c) COMSOL simulation of a single whisker bending.}\label{ml}
  \vspace{-6pt}
\end{figure}

\section{Design and Sensing Principles}

\subsection{Mechanoluminescent Sensing Mechanism}

ML, defined as the emission of light stimulated by mechanical stress, shows great potential for real-time stress sensing applications ~\cite{ML_Overview_zhuang2021mechanoluminescence, ML_Overview_qasem2021recent, ML_Overview_huang2023smart}. 
It offers capabilities such as visualizing stress distribution, remote sensing, dynamic response, and self-powered operation. These capabilities make ML-based stress sensing a highly attractive research area in both academia and industry ~\cite{ML_Application_zhuang2022visualizing, ML_Application_ma2022bimodal, ML_Application_li2023smart}. 
Notably, ZnS:Cu, a type of ML material, displays an exceptionally low threshold for ML initiation, making it a desirable choice for inclusion in sensing mechanisms~\cite{ML_ZnSCu_jeong2014bright, ML_ZnSCu_hou2022interactive}.

\subsubsection{Preliminary Experiments}
The initial set of experiments, displayed in Fig.~\ref{ml}a.ii and iii, utilize sample films composed of ZnS:Cu mixed into a SORTA-Clear$ ^{\mathrm{TM}} $ 37 (SC37). These investigations, conducted in a light-insulated environment, delve deeply into the film stretch ML. To quantify the ML characteristics, we employ a Prtronic flexible electronics tester and a spectrometer, subjecting the ML film samples to tensile and compressive forces. Experiments proceed at a steady speed of  $ 40~\mathrm{mm/s} $, stretching the samples from $ 20~\mathrm{mm} $ to $ 32~\mathrm{mm} $. ML intensity, captured by a NOVA spectrometer, reveals variations with different ZnS:Cu $ wt~\mathrm{\%} $, as seen in Fig.~\ref{ml}a.i. The ML intensity exhibits a monotonic increase when the concentration is amplified from $ 20 wt~\mathrm{\%} $ to $ 60 wt~\mathrm{\%} $. Moreover, an experiment is conducted to confirm the linear relationship between ML intensity and applied force in a film comprising 50 wt\% of the material  (inset of Fig.~\ref{ml}a.i), which shows suitability for dynamic stress sensing. Furthermore, Fig.~\ref{ml}b illustrates the mechanical properties of the film under varying concentrations of ZnS:Cu, tested in compliance with ISO 37 standards using the Universal Testing Machine. It is observed that the mechanical properties undergo a monotonic decrease with an increase in the concentration from $ 20 wt~\mathrm{\%} $ to $ 60 wt~\mathrm{\%} $. This trend suggests a transition in the material's behavior from ductile to brittle as illustrated by the dashed line. After considering both toughness and ML properties, we select a composition of $ 50 wt~\mathrm{\%} $ ZnS:Cu and SC37 as the luminescent layer suitable for tactile sensing applications.

\begin{figure*}[thpb]
  \centering
  \includegraphics[width=1.0\linewidth]{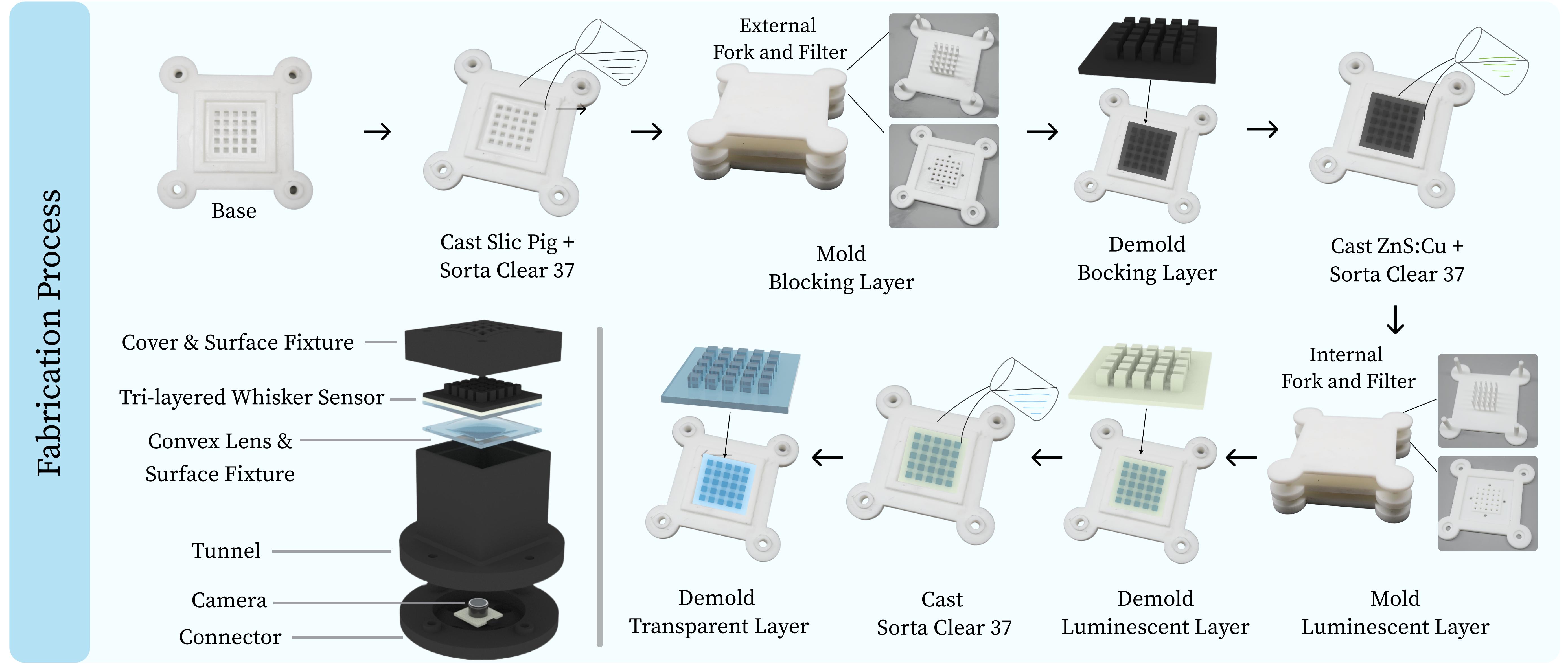}\\
  \caption{The fabrication process of ML whisker elastomer and WSTac.}\label{FabND}
  \vspace{-6pt}
\end{figure*}

\subsection{Tri-Layered Whisker Elastomer Array} 
\subsubsection{Design of the Biomimetic Sensing Elastomer}
Vibrissae (whiskers), equip mammals with an enhanced tactile sense~\cite{Whisker_1,Whisker_2}. The vibrissae’s high sensitivity and rapid response time render them particularly advantageous in object recognition, notably in poorly lit environments. Rodents, for example, primarily rely on vibrissae for navigation and predation~\cite{Whisker_3}.

Our sensor design mimics this animalistic exploration method, utilizing a flexible, biomimetic ML whisker-structured elastomer. In our ML whisker array design, we considered the bending moment of a vibrassae. To be more specific, within our sensor array's whiskers, the forces are transmitted down each whisker column, emulating the transfer along a vibrissae. Sensory mechanoreceptors within the vibrassae follicle convert the movement of the vibrassae shaft into environmental contact data. Similarly, each whisker base within our sensor array emulates these sensory mechanoreceptors, with the photon emission and signal detection. 

The dynamic simulation in Fig.~\ref{ml}c, using COMSOL Multiphysics, explains the ML mechanism in the whisker structure. The model involves an acrylic sphere of $0.5~\mathrm{mm}$ radius sliding across a single $ 1 \times 5~\mathrm{mm}$ SC37 whisker (fixed at the base) at $0.1~\mathrm{m/s}$. The simulation result at $0~\mathrm{s}$, $1~\mathrm{s}$, and $2~\mathrm{s}$ confirms the expected outcome of significant mechanical strain at the base of a single whisker, which is primarily responsible for the emission of ML photons.

\subsubsection{Farbrication of the Elastomer} 
SC37 is employed in all elastomer layers, albeit with minor adjustments for their respective functions as displayed in Table ~\ref{composition-table}. All layers are mixed uniformly in a Planetary Centrifugal Mixer (THINKY MIXER ARE-310).

\begin{table}[thpb]
\caption{Material composition of each elastomer layer.}
\small
\vspace{-8pt}
\label{composition-table}
\begin{center}
\setlength{\tabcolsep}{0.6em}{
\begin{tabular}{l|c}
\toprule
\textbf{Layer} & \textbf{Composition~($ \mathrm{wt} \% $)} \\ 
\hline 
External (Light absorbing) & SC37 (98), Slic Pig (2) \\
Middle (ML) &SC37 (50), ZnS:Cu (50) \\
Innermost (Light transmitting) & SC37 (100) \\
\bottomrule
\end{tabular}}
\end{center}
\vspace{-4pt}
\end{table}

During the fabrication of the ML whisker elastomer, elastomer molding is exploited. As indicated in Fig.~\ref{FabND}, the mold consists of five 3D printed parts: the internal fork and filters, the external fork and filter, and a base mold shaping the elastomer into whisker-like form. The external fork and filter shape the external layer with spaces for the middle layer. Similarly, the internal counterparts allow room for the inner light transmission layer. 

The construction of the ML whisker elastomer sensor includes the following: Firstly, the external layer is transferred onto the base mold and vacuumed at $ 30~\mathrm{kPa} $ to remove bubbles. The external fork is placed onto the mold, initiating the external layer's inner structure, with the filter securing the fork. Upon solidification, the external fork and filter are detached from the base mold while keeping the external layer in its position. The middle layer is then poured on top of the external layer and vacuumed. The internal fork and filter shape the middle layer and are removed after hardening. Subsequently, the innermost layer is added, with existing slots providing shape. No additional forks or filters are needed. Following the solidification of the innermost layer, the ML whisker elastomer is de-molded and ready for system integration. Consequently, the fabrication process is cost-efficient, amounting to 12 USD.

\subsection{Mechanical Design}
The WSTac system comprises five key components: the cover, the ML whisker elastomer, a convex lens, a tunnel, and a camera.

The cover secures the ML whisker elastomer at the top of the tunnel by sandwiching it between itself and a convex lens. Shared M3 screw holes connect these components, enabling easy part replacement and facilitating task-specific elastomer incorporation.  

Both the convex lens and the outer shell boast a curvature of 30\degree, a design choice intended to maximize the whiskers' exposure to varying deflection directions and optimize their contact positioning. The photon generation process within the ML whisker elastomer is shielded from environmental influences as the photons are channeled through the protective tunnel, ensuring the camera captures photon signals without significant signal loss.  

Secured at the tunnel base using M3 screws,  is a custom made low-cost 30~fps camera module with a resolution of $ 1280 \times 720 $. A connector affixed to the base of the tunnel allows for the WSTac system to be attached to a robotic arm. 

\subsection{Algorithms}
WSTac’s software component effectively identifies and extracts tactile signals from 25 taxels (ML whiskers) with camera imaging. It is achieved through pin identification in real-time tactile imagery with a resolution of $ 400 \times 400 $. The real-time pin identification recognizes taxels equivalent to 50-unit squares. 
From Fig.~\ref{ml}a given that taxels emit mainly green light, 
we generate time-series signals
$ {\mathbf{O} (t)}\!=\!\{ o_{i,j} (t) \}_{i,j=1,2,\dots,5} $ by calculating the mean of the green channel for all pixels within each taxel and normalizing each by 255,
where $ i $ signifies row and $ j $ denotes column.

\begin{figure}[thpb]
  \centering
  \includegraphics[width=0.9\linewidth]{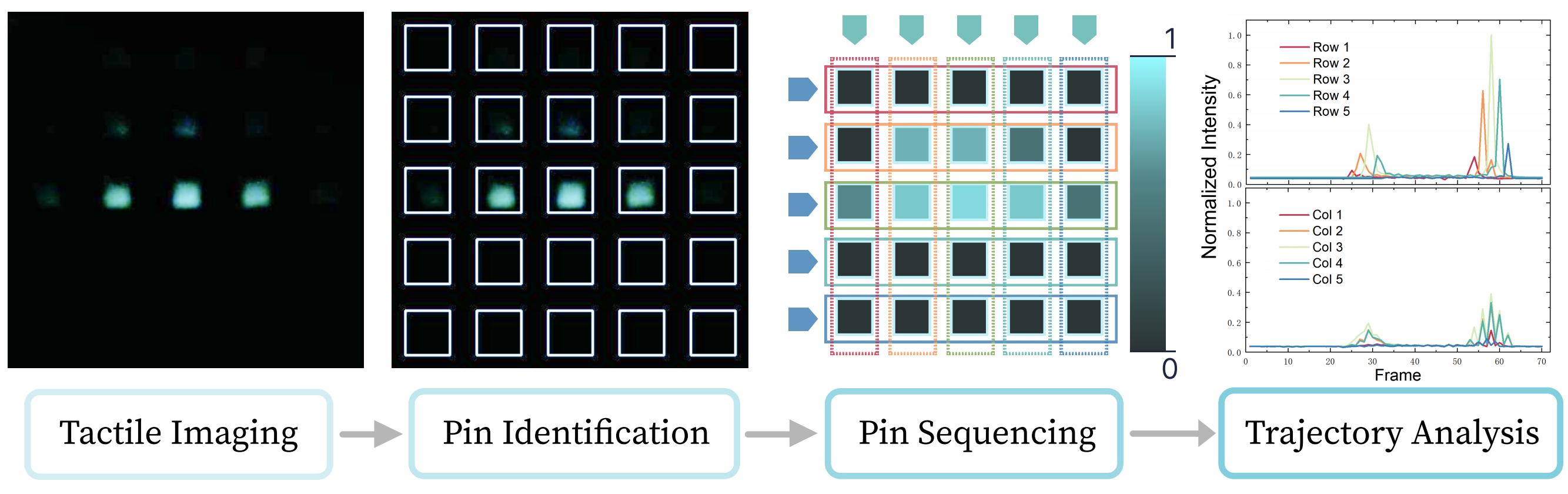}\\
  \caption{Feature extraction pipeline of WSTac. (From  tactile images to multivariate time series. ) }
  \label{pipeline}
  \vspace{-6pt}
\end{figure}

In feature extraction, we aim to minimize the dimensions of multivariate time series, while differentiating four key parameters: direction, speed, and the pattern and depth of the surface. 
Our method is inspired by the electronic tactile sensor array's functioning~\cite{Algorithms_electronic_1}, which uses a row+column electrode structure to eliminate crosstalk and reduce the number of connection wires from $i \times j$ to $i + j$. We adopt a similar approach for VTBS to decrease the algorithm complexity from $\Theta(i \times j)$ to $\Theta(i + j)$.
Regardless of whether WSTac slides vertically or horizontally, the directionally-relevant sums of $ {\mathbf{O} (t)} $ allow us to identify the speed, specimen's pattern and depth. Thus feature signals $ {\mathbf{F} (t)}\!=\!\{ f_{k} (t) \}_{k=1,2,\dots,10} $ can be calculated as:

\begin{equation}
{\mathbf{f}_k} (t) = \begin{cases}
\ln\left(~\sum\limits^{5}_{j=1}  o_{k,j} (t) ~ \right) ,\quad & k = 1,2,3,4,5 \\
\\
\ln\left(~\sum\limits^{5}_{i=1}  o_{i,k} (t) ~ \right) ,\quad & k = 6,7,8,9,10
\end{cases} 
\vspace{10pt}
\label{eqFeature1}
\end{equation}

These features not only contain the sliding direction information but also reduce the original 25 time-series to 10. Fig.~\ref{pipeline} exhibits WSTac's event-driven tactile feature extraction process.

Furthermore, for enhancing data utilization, based on sliding windows method, WSTac can autonomously detect valid tactile signals.  
The camera captures tactile images for each frame $ t $.  From these images,  $ {\mathbf{F} (t)} $ is extracted by Eqn. (\ref{eqFeature1}). 
Prior to the WSTac making contact with the surface, we compute the sum values within a sliding window and get the average values across the first five windows for each feature signal  $ \eta_{k} $.
While the WSTac remains active, if any of the $ {f_k} (t) $ meets the trigger condition, Eqn. (\ref{eqFeature2}), tactile sample signals $ {\mathbf{X}} \in \mathbb{R}^{10 \times l} $ of fixed length $ l = 70 $ can be obtained for application or further processing. To avoid resampling the same signal, the sliding window is suppressed within $ l $ frames. 

\begin{algorithm}[!htp]
  \caption{Event-Driven Tactile Data Collection Method}
  \label{alg:myalgorithm}
\begin{algorithmic}[1]
\REQUIRE Feature signals $ {\mathbf{F} (t)} $ - per frame $ t $ from Eqn. (\ref{eqFeature1});
\STATE WSTac initialization $ t = 0 $
\FOR{$k\leftarrow 1$ \TO $10$}
\STATE $ \eta_{k} =  \sum^{5 \cdot m-1}_{w=0}  f_{k} (t + w) / 5$
\ENDFOR
\STATE $ t \leftarrow t + 5 \cdot m $
\WHILE{WSTac is active}
\FOR{$k\leftarrow 1$ \TO $10$}
\IF{
\begin{equation}  \sum^{m-1}_{w=0}  f_{k} (t + w) > b \cdot \eta_{k} 
\label{eqFeature2}
\end{equation}}
\STATE {${\mathbf{X}}   \leftarrow [ {\mathbf{F}} (t - c),\dots,{\mathbf{F}} (t  + l - c  - 1 ) \} ] $} 
\STATE 3 tactile perception tasks \OR dataset ${ \leftarrow \mathbf{X}}$
\STATE $ t \leftarrow t + l $
\STATE \textbf{break for}
\ENDIF
\STATE $ t \leftarrow t + m $
\ENDFOR
\ENDWHILE
\end{algorithmic}
\end{algorithm}

{\textbf{Algorithm~\ref{alg:myalgorithm}}} describes the event-driven tactile data collection process. Here, 
$ m $ is the width of sliding window,
$ c $  refers to the number of frames to backtrack and $ b $ is the threshold trigger multiplier. 

\section{Experiment and Result}

\begin{figure}[!thpb]
  \centering
  \includegraphics[width=1.0\linewidth]{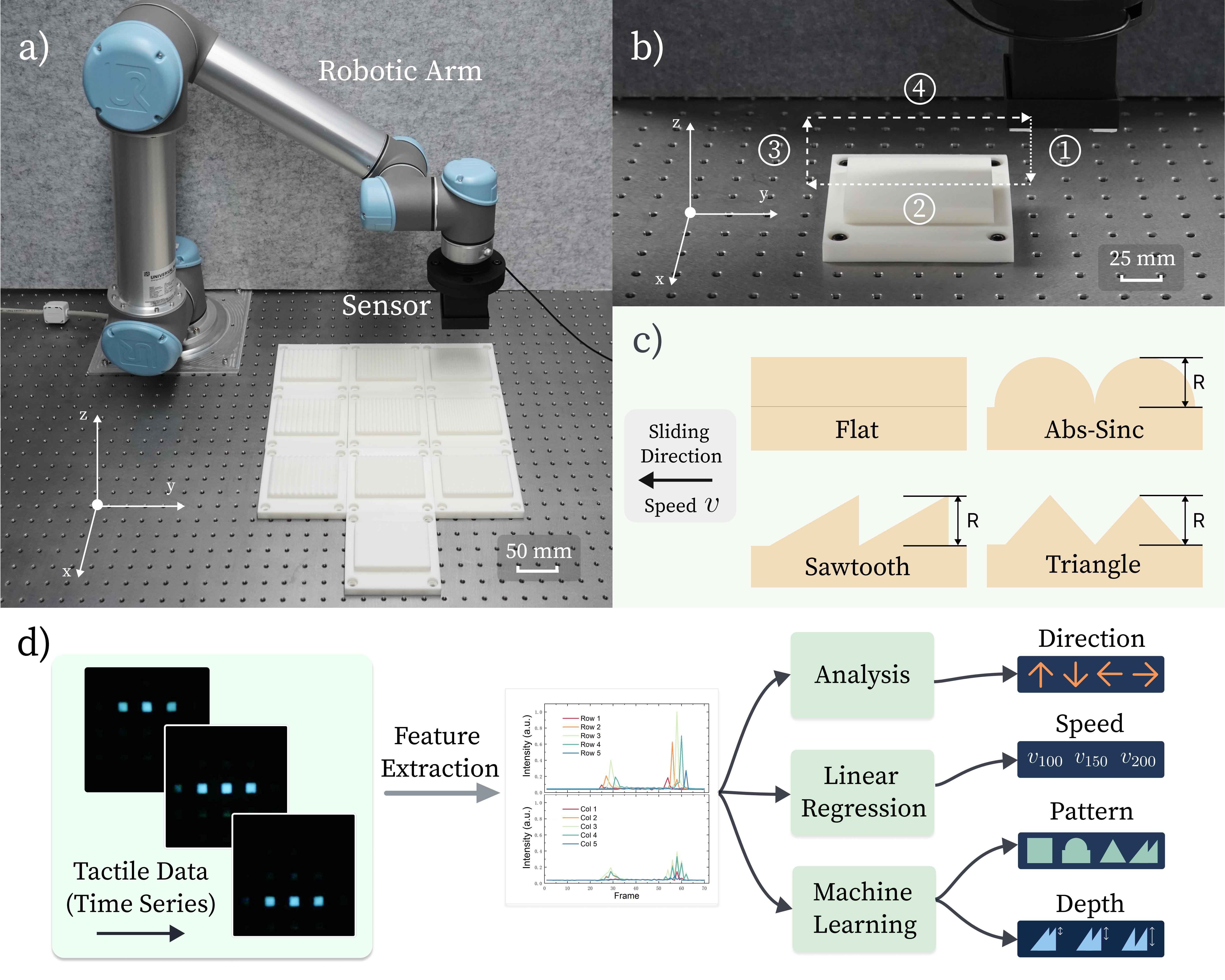}\\
  \caption{Experimental setup: (a)WSTac attached to a 6-dof industrial robot arm (UR5) and 10 3d-printed textures; (b) The sliding path for experiments;  (c) Design of experiment specimens; (d) Data processing pipeline for application. }\label{UR5}
\end{figure}

We establish an experimental platform to study WSTac's sliding behaviour on diverse 3D printed specimens under varying conditions. 
Fig.~\ref{UR5}a illustrates the complete setup, featuring a six degrees of freedom robotic arm (UR5) with a WSTac, and ten textured specimens. 
As shown in Fig.~\ref{UR5}b, the robotic arm collects data along a fixed trajectory to perform the following three tasks.

\subsection{Speed Regression}
We determine the sliding speed of WSTac by analyzing the total luminescence duration. $  {\mathbf{X}} $ from eleven different sliding velocities $ \mathbf{V} \!=\! \{ 100~\mathrm{mm/s}, 110~\mathrm{mm/s},\dots, 200~\mathrm{mm/s} \}$ are analyzed, with five samples each, along a predetermined trajectory.

\begin{figure*}[thpb]
  \centering
  \includegraphics[width=1.0\linewidth]{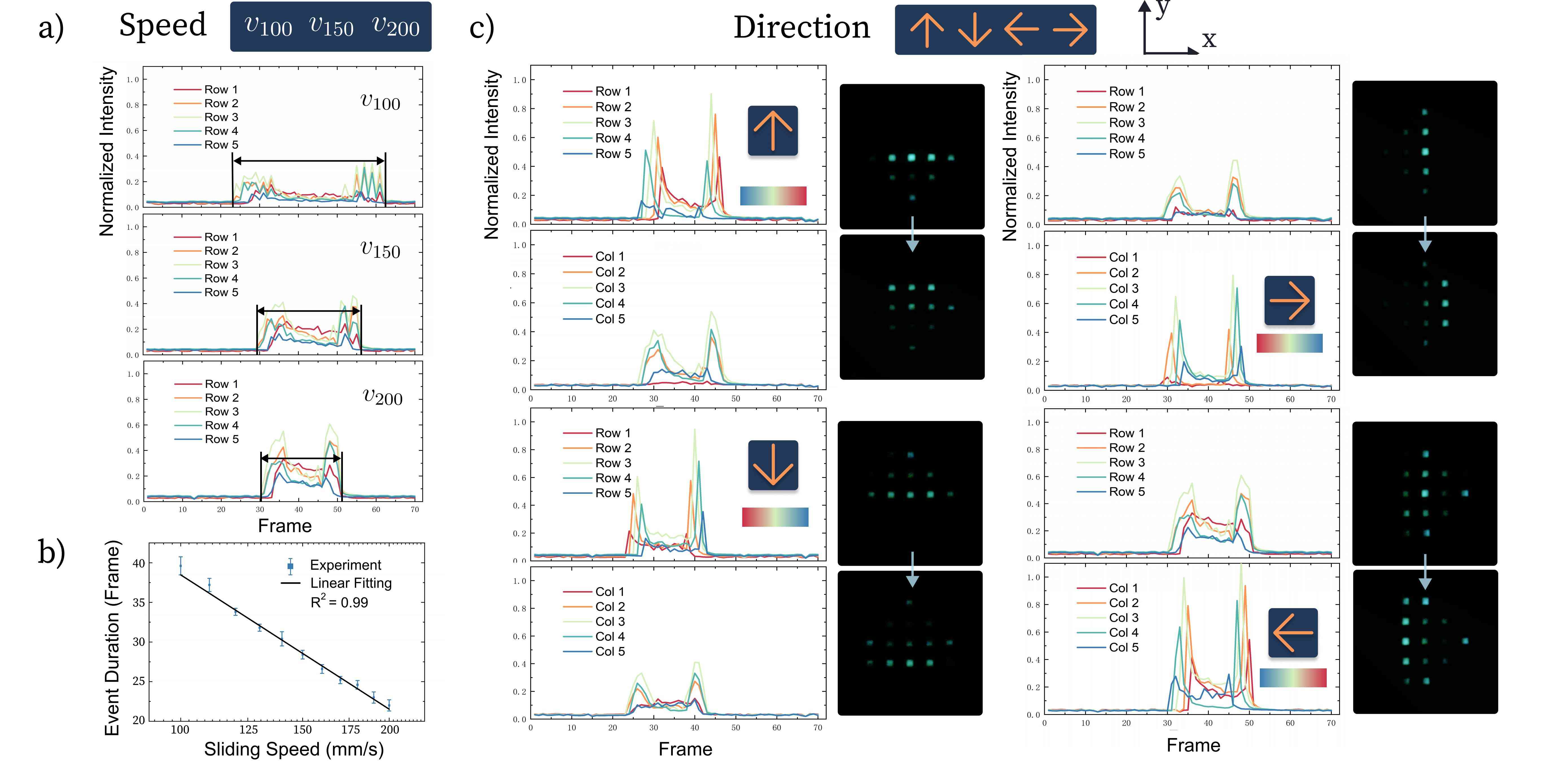}\\
  \caption{WSTac's Data analysis: (a) Total luminescence duration for sliding speeds $ \mathbf{V}\!=\! \{ 100~\mathrm{mm/s}, 150~\mathrm{mm/s}, 200~\mathrm{mm/s} \}$; (b) Negative correlation between Event Duration and Sliding speed; (c) Visualization of data from four sliding directions $ \mathbf{D}\!=\! \{ 0^\mathrm{\circ}, 90^\mathrm{\circ}, 180^\mathrm{\circ}, 270^\mathrm{\circ} \}$ . }
  \label{data}
  \vspace{-6pt}
\end{figure*}

The total luminescence duration for each speed is obtained by first employing a threshold value of 0.0475 to distinguish valid from invalid frames within the signal. Frames exceeding the threshold are considered valid. Within these valid frames, a non-zero sum of signal values at each frame $t$ indicates valid data. The signal start and end times are identified as the first and last occurrences of valid data respectively, as shown in Fig.~\ref{data}a. The total luminescence duration (Event Duration) is subsequently calculated by subtracting the start time from the end time.

Our study, as shown in Fig.~\ref{data}b, reveals a negative correlation between
sliding speed and event duration, represented by the logarithmic regression model $ y \!=\! 151.06 - 56.29 ~ log(x) $. The slope of $ -56.29 $ signifies that higher Sliding Speed leads to a proportional decrease in Event Duration, supported by our model’s  strong coefficient of determination value of 0.99.

\subsection{Directional Identification}
Data from four sliding directions $ \mathbf{D}\!=\! \{ 0^\mathrm{\circ}, 90^\mathrm{\circ}, 180^\mathrm{\circ}, 270^\mathrm{\circ} \}$ is depicted in Fig.~\ref{data}c, visualized in either sequential or reverse order within rows or columns. The luminescence sequence distinguishes between $+ \vec{y}$ and $- \vec{y}$ on the y-axis and $+ \vec{x}$ and $- \vec{x}$ on the x-axis. $+ \vec{y}$ is represented by a reverse order in the row list, while $- \vec{y}$ is represented by a sequential order. Similarly, $+ \vec{x}$ corresponds to a sequential order in the column list, while $- \vec{x}$ corresponds to a reverse order. Ultimately, these variations in luminescence sequence successfully determine distinct sliding directions.

\subsection{Texture Classification}
For texture recognition, we design ten specimens with uniform geometrical dimensions and depths on their contact surfaces. Fig.~\ref{UR5}c shows their patterns, inspired by common waveforms including flat, abs-sinc, sawtooth, triangle ($ \mathbf{P}\!=\! \{ \mathrm{Fla}, \mathrm{Sin}, \mathrm{Saw}, \mathrm{Tri} \}$). For non-flat patterns, we set the texture depth ($ \mathbf{R}\!=\! \{ 0~\mathrm{mm}, 2~\mathrm{mm}, 3~\mathrm{mm}, 4~\mathrm{mm} \}$) as half the periodic lengths, with $ \mathbf{R}\!=\! 0~\mathrm{mm} $ corresponding to $ \mathbf{P}\!=\! \mathrm{Fla} $. 

The robotic arm slides along a preset trajectory 100 times for each specimen.
Each sliding generates a tactile signal $ {\mathbf{X}} $ of fixed length, which is automatically assigned corresponding labels ${\mathbf{Y}}$. The collected  $ {\mathbf{X}} $ data from each specimen is compiled into a dataset, which can be used to train machine learning models for predicting the ten 3D printed specimens, pattern ~$ \mathbf{P}$, and depth ~$ \mathbf{R}$. 

To showcase WSTac's outstanding performance and facilitate real-time deployment of the model, we utilize three classic machine learning models: Linear-SVM~\cite{model_SVM}, Random Forest~\cite{model_Random_forests} , and Xgboost~\cite{model_Xgboost}. The dataset is split into training and test sets at a 9:1 ratio. We build separate models for the three classification tasks (10 specimens, patterns $ \mathbf{P}$, and depths $ \mathbf{R}$).

Test accuracy is used as a metric to evaluate the models. 
Table~\ref{Results-table} shows that high-precision classification could be achieved for each task using machine learning models, negating the necessity for computationally expensive deep learning networks. This validates the effectiveness of our sensor design and feature extraction algorithms in harvesting high-quality dynamic tactile information. 

\begin{table}[thpb]
\caption{Texture classification results in 100 test samples.}
\small
\vspace{-8pt}
\label{Results-table}
\begin{center}
\setlength{\tabcolsep}{0.6em}{
\begin{tabular}{l|c|c|c}
\toprule
\textbf{Label~(Num of class)} & \textbf{Linear-SVM} & \textbf{Random~forest} & \textbf{Xgboost}   \\ 
\hline 
Specimens~(10) & 83\%  & 94\% &  95\%   \\
Patterns~(4)  &  91\%  & 98\% &  97\% \\
Depths~(4)  & 92\%  & 99\% & 98\%  \\
\bottomrule
\end{tabular}}
\end{center}
\vspace{-4pt}
\end{table}

\section{Conclusions and Discussions}
In this study, we demonstrate WSTac: a VBTS that does not require LED modules, yet successfully replicates the fundamental functionalities of a typical VBTS. Experiments show that WSTac exhibits impressive precision across three tactile tasks.

Although this work focuses on showcasing the basic sensory capabilities of VBTS, we believe it is of significant importance to the research in this field. WSTac represents an experimental foray outside the traditional VBTS paradigm. Our vision for the future involves integrating WSTac into robotic grippers or human-machine interfaces, coupled with advanced algorithms, to potentially surpass the performance of LED-equipped VBTS in certain applications. Enhancements could include the integration of a charge-coupled device, a high-speed camera, or boosting the ML properties.

\appendix
\setcounter{secnumdepth}{0}
\bibliographystyle{ACM-Reference-Format}
\bibliography{sample-base}

\end{document}